\documentclass[11pt, a4paper]{article}

\usepackage[margin=1in]{geometry}
\usepackage{graphicx}
\usepackage{booktabs}
\usepackage{hyperref}
\usepackage{float}
\usepackage{subcaption}
\usepackage{cite}
\usepackage{siunitx}
\usepackage{algorithm}
\usepackage{algorithmic}
\usepackage{amsmath}
\usepackage{amssymb}
\usepackage{authblk}

\usepackage{fancyhdr}
\title{\textbf{FLNet: Flood-Induced Agriculture Damage Assessment using Super Resolution of Satellite Images}}

\author[1]{Sanidhya Ghosal\thanks{Corresponding author: sanidhyakumarghosal@gmail.com}}
\author[2]{Anurag Sharma}
\author[3]{Sushil Ghildiyal}
\author[3]{Mukesh Saini}

\affil[1]{Annam.AI CoE, MoE, Indian Institute of Technology Ropar, Punjab, India}
\affil[2]{Department of Mathematical Sciences, Rajiv Gandhi Institute of Petroleum Technology, Jais, UP, India}
\affil[3]{Department of Computer Science and Engineering, Indian Institute of Technology Ropar, Punjab, India}

\date{}

\begin{document}

\maketitle

\begin{abstract}
Distributing government relief efforts after a flood is challenging. In India, the crops are widely affected by floods; therefore, making rapid and accurate crop damage assessment is crucial for effective post-disaster agricultural management. Traditional manual surveys are slow and biased, while current satellite-based methods face challenges like cloud cover and low spatial resolution. Therefore, to bridge this gap, this paper introduced FLNet, a novel deep learning based architecture that used super-resolution to enhance the 10 m spatial resolution of Sentinel-2 satellite images into 3 m resolution before classifying damage. We tested our model on the Bihar Flood Impacted Croplands Dataset (BFCD-22), and the results showed an improved critical ``Full Damage" F1-score from 0.83 to 0.89, nearly matching the 0.89 score of commercial high-resolution imagery. This work presented a cost-effective and scalable solution, paving the way for a nationwide shift from manual to automated, high-fidelity damage assessment.

\vspace{0.5cm}
\noindent \textbf{Keywords:} Flood Damage Assessment, Super Resolution, Sentinel-2, U\textsc{Net}, NDVI, Remote Sensing.
\end{abstract}

\section{Introduction}

Floods are among the most frequent and devastating natural disasters, posing a significant and growing risk to agriculture and rural communities, particularly in agrarian societies like India \cite{IPCCAR6SPM, IPCCAR6Ch11, UNDRR2020}. In the aftermath of a flood, a fast and objective assessment of crop damage is the first critical step for government agencies and insurance bodies to deliver timely relief and compensation. While traditional methods relied on manual, on-the-ground surveys, this approach was often slow, expensive, prone to human bias, and impractical for covering large affected areas.

Remote sensing, using satellite imagery, offers a powerful alternative for wide-area, with highly generalizable models. For decades, analysts have used optical indices like the Normalized Difference Vegetation Index (NDVI) to monitor crop health. By comparing pre-flood and post-flood NDVI imagery a method known as temporal differencing ($\Delta$NDVI) we could get a direct signal of vegetation loss and crop failure \cite{Tucker1979, Rouse1974, Singh1989, Coppin1996}. While the Normalized Difference Water Index (NDWI) can effectively map the surface area covered by floodwater, its main limitation is that it cannot quantify the health of the crops themselves. It doesn't measure damage to plants that are stressed, waterlogged, or damaged but not fully submerged. Furthermore, Synthetic Aperture Radar (SAR) could provide crucial data by imaging through the cloud cover typical of flood events \cite{Tarpanelli2022}.

However, applying these methods effectively in regions dominated by small farms presented a significant challenge. The freely available Copernicus Sentinel-2 satellite constellation provides valuable multi spectral images at a 10-meter spatial resolution \cite{Drusch2012}. While useful, this resolution is often too coarse for the narrow, fragmented agricultural landscapes found in much of India. A single 10m pixel can often blend signals from multiple different farm plots, weakening the true damage signal and blurring boundaries. This well-known issue is called the ``mixed-pixel problem" \cite{Keshava2002, Bioucas2012}. During monsoon floods, cloud and shadow cover were also major issues, and the process of masking these areas could reduce data availability and introduce sampling bias \cite{MainKnorn2017}. While high-resolution commercial imagery from constellations like PlanetScope (3-meter resolution) can overcome the resolution issue, its high cost and different sensor characteristics complicated its use and made it unfeasible for routine, large-scale deployment \cite{Houborg2018, Sadeh2021}.

To address this challenge, this paper proposed a novel deep learning pipeline, FLNet (Flood Loss Damage Classification Network), that bridged this resolution gap. We treated single-image super-resolution (SISR) as a ``virtual sensor" that could generate high-resolution detail from free Sentinel-2 data. Our approach used an Enhanced Deep Super-Resolution (EDSR) model \cite{Lim2017} to learn a direct mapping from a 10m Sentinel-2 NDVI image to a 3m NDVI equivalent. Crucially, we trained this model using real 3m Planet Scope NDVI images as the ground truth, avoiding biasses that can arise from synthetic training data \cite{Qi2025RSISR}. After super-resolving both the pre- and post-flood NDVI images, we computed the $\Delta$NDVI at 3m and fed this feature into a U\textsc{Net} segmentation model \cite{Ronneberger2015}, an architecture representative of modern deep models that have excelled at this type of pixel-level classification task \cite{Zhu2017}. We tested this pipeline on the October 2022 flood in Muzaffarpur, Bihar, and found that the super-resolution step significantly sharpened field boundaries and materially improved the detection of the critical ``Full Damage" class compared to using native 10m inputs.

The main contributions of this paper are as follows:
\begin{itemize}
    \item The proposal of FLNet, a novel, end-to-end architecture that coupled EDSR-based super-resolution with a $\Delta$NDVI-driven U\textsc{Net} segmentation model. This pipeline was explicitly designed to target smallholder farms where the mixed-pixel effect undermined standard change detection methods \cite{Lim2017, Qi2025RSISR, Ronneberger2015, Zhu2017}.
    \item The development and curation of the Bihar Flood Impacted Croplands Dataset (BFCD-22). This new dataset comprises co-registered pre- and post-flood Sentinel-2 and PlanetScope imagery over Muzaffarpur and includes quality masks and damage labels to enable a fair and reproducible comparison between different mapping methods.
\end{itemize}

\section{Related Work}
Our work built upon established research in several key areas. We first reviewed the standard remote sensing methods for assessing flood damage in agriculture. We then discussed the deep learning models commonly used for image segmentation, followed by an overview of super-resolution techniques in remote sensing. Finally, we touched upon two complementary technologies: Synthetic Aperture Radar (SAR) for its all-weather capabilities, and modern approaches to learned cloud removal. By reviewing these fields, we could better position our specific design choices and highlight our contribution.

\subsection{Assessing Agricultural Flood Damage with Remote Sensing}
Prior research established the Normalized Difference Vegetation Index (NDVI) as a reliable indicator of vegetation health \cite{Tucker1979, Rouse1974}, and showed that calculating its change over time ($\Delta$NDVI) is an effective way to quantify crop stress after a event\cite{Singh1989,Coppin1996}. Following this, deep learning models like the U\textsc{Net} architecture have proven highly effective for precise, pixel-wise classification of such remote sensing data \cite{Zhu2017, Ronneberger2015}. The U\textsc{Net}'s ability to preserve fine boundary details is especially valuable for this task \cite{Maggiori2017}. This work inspired our choice to use $\Delta$NDVI as the core feature and U\textsc{Net} as the segmentation model. However, a shared limitation of these methods is their dependence on the spatial resolution of the satellite imagery. Both the index calculation and the model's performance lead to inaccurate results in areas with small farms when using blurry, low-resolution data.

Research has demonstrated that Synthetic Aperture Radar (SAR) is an invaluable all-weather tool for flood monitoring, as it can penetrate clouds that block optical sensors, thus providing more imaging opportunities during critical flood periods \cite{Tarpanelli2022}. This inspired us to consider SAR as a powerful complementary data source for future enhancements to our pipeline. However, SAR primarily measures surface texture and moisture, and does not provide a direct signal of vegetation health in the same way NDVI does, making it less suitable as a standalone feature for our specific goal of assessing crop stress.

\subsection{Super-Resolution in Remote Sensing}
Studies in super-resolution have established strong baseline models like EDSR \cite{Lim2017} and have recommended training with real high-resolution images as supervision instead of synthetic data for better real-world performance \cite{Qi2025RSISR}. These findings directly inspired our core method of using an EDSR model trained on real 3m PlanetScope imagery. The primary limitation of existing work is that most remote sensing super-resolution has focused on creating visually appealing RGB images, with fewer studies applying it directly to scientific indices like NDVI to improve a downstream analytical task like damage classification.

\subsection{Learned Cloud Removal Techniques}
Recent advancements have produced AI models that can attempt to digitally reconstruct the clear view of the ground underneath clouds in satellite imagery \cite{Ghildiyal2025JSTARS, Ghildiyal2022MIPR, Ghildiyal2024EJA}. This work inspired our future plans to integrate such techniques to increase the amount of usable data. The main limitation of these methods is the risk of altering the image's precise pixel values (its radiometry), which could introduce errors into the sensitive mathematical calculation of $\Delta$NDVI and lead to false damage detection.

\section{Methodology}
\begin{figure}[!t]
    \centering
    \includegraphics[width=0.95\textwidth]{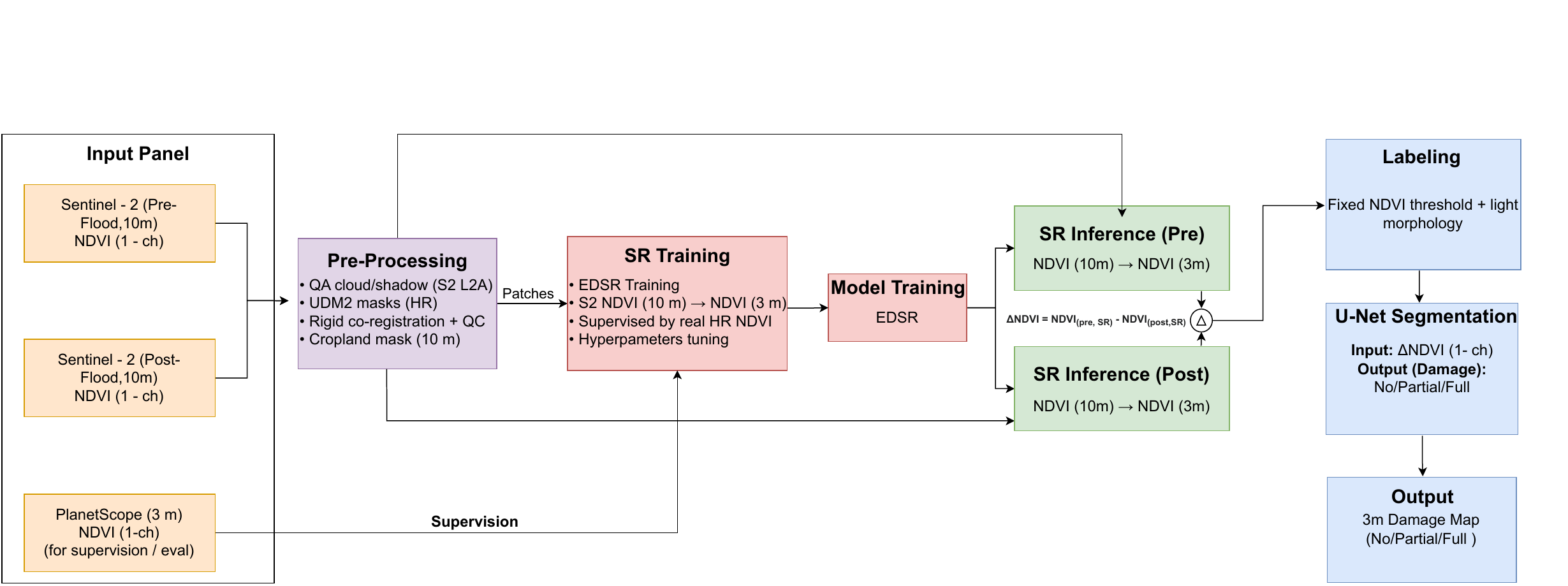}
    \caption{Illustrates the architecture of the FLNet model.}
    \label{fig:flowchart}
\end{figure}

This section details the technical pipeline of our FLNet model, which was designed to map flood-induced agricultural damage using freely available Sentinel-2 data. The entire workflow is summarized in Figure~\ref{fig:flowchart}. The process began with acquiring and preparing the satellite data, which then moved through three main stages: (1) an EDSR-based super-resolution module enhanced the pre- and post-flood NDVI images; (2) a change feature ($\Delta$NDVI) was generated from these enhanced images; and (3) a U\textsc{Net}-based segmentation model classified the damage. Throughout this process, we used careful quality masking and image co-registration to ensure the reliability of our results, and we trained our super-resolution model using real 3m PlanetScope imagery as the ground truth.

\subsection{Motivation}
In agricultural landscapes with many small farms, a single 10m pixel from Sentinel-2 often covers multiple fields with different crop conditions. This ``mixed-pixel problem" weakened the signal of any changes and blurred the boundaries between properties. During monsoon floods, this issue was made worse by clouds and shadows, which reduced the amount of usable data. While high-resolution 3m imagery from PlanetScope solved the mixed-pixel problem, it can be costly and has fewer spectral bands than Sentinel-2. Our goal was to overcome these challenges by using single-image super-resolution directly on the NDVI values. This approach acted as a ``virtual sensor", allowing us to achieve the sharp detail of high-resolution imagery while still using the free and spectrally rich data from Sentinel-2.

\subsection{Problem Formulation}
To formally define our method, we first specified our inputs and outputs. We represented the pre-flood Sentinel-2 (S2) NDVI image at 10m resolution as $I_{\mathrm{pre}}^{\mathrm{S2}}$ and the post-flood Sentinel-2 (S2) NDVI image as $I_{\mathrm{post}}^{\mathrm{S2}}$. Both are defined as $I_{\mathrm{pre}}^{\mathrm{S2}}, I_{\mathrm{post}}^{\mathrm{S2}} \in \mathbb{R}^{H\times W}$, where $\mathbb{R}^{H\times W}$ denotes a two-dimensional matrix of real numbers with height $H$ and width $W$. The corresponding high-resolution PlanetScope (PS) NDVI image at 3m, which serves as our ground truth, is denoted as $I^{\mathrm{PS}} \in \mathbb{R}^{\hat H\times \hat W}$, where $\mathbb{R}^{\hat H\times \hat W}$ represents the higher-resolution matrix with dimensions $\hat H \times \hat W$.

Our first goal was to learn a super-resolution mapping function, denoted as $\mathcal{S}_\theta$, where $\theta$ represents the learnable parameters of the (EDSR) model. This function $\mathcal{S}_\theta: \mathbb{R}^{H\times W}\rightarrow \mathbb{R}^{\hat H\times \hat W}$ is trained to map an image from the low-resolution space ($H \times W$) to the high-resolution space ($\hat H \times \hat W$). This function, when applied to our images, produced the super-resolved outputs: $\hat I_{\mathrm{pre}}^{\mathrm{S2}}=\mathcal{S}_\theta(I_{\mathrm{pre}}^{\mathrm{S2}})$ and $\hat I_{\mathrm{post}}^{\mathrm{S2}}=\mathcal{S}_\theta(I_{\mathrm{post}}^{\mathrm{S2}})$, where $\hat I_{\mathrm{pre}}^{\mathrm{S2}}$ and $\hat I_{\mathrm{post}}^{\mathrm{S2}}$ are the generated high-resolution versions of the pre- and post-flood images, respectively. We trained this model by supervising it with the real PlanetScope data, $I^{\mathrm{PS}}$, using an $\ell_1$ reconstruction loss (which measures the mean absolute error between the generated output and the ground truth) to minimize the error.

Further, we used these high-resolution outputs to create a single-channel change feature, $\Delta\text{NDVI}$, calculated as the difference between the super-resolved pre- and post-flood images: $\Delta\text{NDVI}=\hat I_{\mathrm{pre}}^{\mathrm{S2}}-\hat I_{\mathrm{post}}^{\mathrm{S2}}$. This resulting $\Delta\text{NDVI}$ is a change map in the high-resolution space $\mathbb{R}^{\hat H\times \hat W}$. Finally, we learned a segmentation mapping function, $\mathcal{G}_\phi$, where $\phi$ represents the learnable parameters of the (UNET) model. This function $\mathcal{G}_\phi:\mathbb{R}^{\hat H\times \hat W}\rightarrow \{0,1,2\}^{\hat H\times \hat W}$ takes the $\Delta$NDVI feature map as input and classifies each high-resolution pixel into an integer label representing one of three categories: \{0: No, 1: Partial, 2: Full\} damage. This was trained using a cross-entropy objective. The combined end-to-end model, $(\mathcal{S}_\theta,\mathcal{G}_\phi)$, therefore acted solely on NDVI values, with the super-resolution component ($\mathcal{S}_\theta$) being trained with real 3m data to avoid the artifacts common with synthetic downsampling.

\subsection{Network Architecture}
The FLNet architecture combined a super-resolution backbone with a segmentation head in a simple and effective design. For clarity, we summarized the three main modules here before detailing them in the following subsections.
\begin{enumerate}
    \item The EDSR Super-Resolution Module: This module learned to upscale 10m NDVI images to 3m. It was trained using pairs of co-registered Sentinel-2 and PlanetScope images with an $\ell_1$ loss (detailed in Section~\ref{sec:sr-edsr}).
    \item The Change Feature Module: This module took the super-resolved pre- and post-flood images and computed the change between them ($\Delta$NDVI). It was also used to generate the ground-truth labels for training by applying fixed thresholds to the $\Delta$NDVI values (detailed in Section~\ref{sec:delta-label}).
    \item The U\textsc{Net} Segmentation Module: This module took the $\Delta$NDVI map as input and classified each pixel into \{No, Partial, Full\} damage categories using a cross-entropy loss (detailed in Section~\ref{sec:unet}).
\end{enumerate}
Our main design choices were to perform super-resolution directly on the NDVI values, to supervise this process with real 3m imagery from PlanetScope, and to couple the result with a simple, single-channel U\textsc{Net}. These choices emphasized reliability, simplicity, and efficiency.

\subsection{Super-Resolution with EDSR}\label{sec:sr-edsr}
The core of our super-resolution stage was the EDSR model \cite{Lim2017}, which we configured to work with single-band NDVI data. The model used sixteen residual blocks and sixty-four feature channels, with an upscaling factor designed to convert 10m inputs to 3m outputs. To train it, we extracted aligned patches from our co-registered Sentinel-2 and PlanetScope image pairs. The low-resolution 10m Sentinel-2 NDVI served as the input, and the high-resolution 3m PlanetScope NDVI for the same time and location served as the target. We optimized the model using an $\ell_1$ reconstruction loss with the Adam optimizer and learning rate scheduling. As recommended by recent remote sensing research, supervising the model with real high-resolution imagery helped it learn more realistic details and mitigated issues that arise with synthetic training data \cite{Qi2025RSISR}.

\subsection{Damage Feature Generation and Labeling}\label{sec:delta-label}
After the super-resolution stage, we generated the high-resolution pre- and post-flood NDVI images, which we denoted as $\mathrm{NDVI}_{\text{pre, SR}}$ and $\mathrm{NDVI}_{\text{post, SR}}$. The change feature was then calculated as a simple subtraction:
\begin{equation}
\Delta \text{NDVI}\;=\; \mathrm{NDVI}_{\text{pre, SR}} \;-\; \mathrm{NDVI}_{\text{post, SR}}
\end{equation}
In this feature map, larger positive values indicate a significant loss of vegetation, which is consistent with flood-induced crop damage \cite{Singh1989,Coppin1996}. We also used this $\Delta$NDVI map to generate the ground-truth labels for training our segmentation model. This was done by applying a set of fixed thresholds to the pixel values, which separated them into the \{No, Partial, Full\} damage classes. A light morphological smoothing was then applied to the resulting label map to ensure spatial consistency.

\subsection{U\textsc{Net}-based Damage Segmentation}\label{sec:unet}
The final stage of our pipeline used a U\textsc{Net} model \cite{Ronneberger2015} to perform the damage classification. The network had four encoder-decoder stages with standard skip connections. It took the single-channel $\Delta$NDVI map as input and outputted a three-class map corresponding to our damage categories. We trained the model using a cross-entropy objective, with the option to use focal loss or class weighting to handle the imbalance between the number of damaged and undamaged pixels. Finally, a simple post-processing step involving small object removal was applied to the output to reduce "salt-and-pepper" noise near the boundaries of farm plots.

\begin{algorithm}[H]
\caption{FLNet: super resolution aided $\Delta$NDVI damage mapping}
\label{alg:flnet}
\begin{algorithmic}[1]
\REQUIRE Sentinel-2 L2A pre/post scenes; PlanetScope 3\,m scenes (for supervision/eval); cropland mask
\STATE Preprocess by masking clouds/shadows (S2 L2A, PS UDM2), orthorectify and co register S2/PS, compute NDVI for all dates, and align S2 to the 3\,m grid.
\STATE Extract aligned NDVI patches (e.g., $256{\times}256$ at 3\,m) within cropland.
\STATE Train EDSR by minimizing $\ell_1(\mathcal{S}_\theta(I_{\mathrm{S2}}), I_{\mathrm{PS}})$ with Adam and early stopping on validation.
\STATE Run SR inference to produce $\hat I_{\mathrm{pre}}^{\mathrm{S2}}$ and $\hat I_{\mathrm{post}}^{\mathrm{S2}}$ at 3\,m.
\STATE Compute $\Delta$NDVI=$\hat I_{\mathrm{pre}}^{\mathrm{S2}}-\hat I_{\mathrm{post}}^{\mathrm{S2}}$ and derive labels via fixed thresholds with light morphology.
\STATE Train U\textsc{Net} on $\Delta$NDVI to predict \{No, Partial, Full\}, with early stopping and optional class weights.
\STATE Inference produces 3\,m damage maps; evaluate SR with PSNR/SSIM vs.\ PlanetScope and segmentation with classwise F1.
\end{algorithmic}
\end{algorithm}

\section{Experiments and Results}\label{sec:experiments}
This section presented the detailed experiments conducted to validate our FLNet model. We began by describing our study area and the custom dataset we created, the Bihar Flood-Impacted Croplands Dataset (BFCD-22), along with our data preprocessing workflow. We then outlined the implementation details and training setup for both the super-resolution and segmentation models. Following this, we defined the metrics used for evaluation and presented the quantitative and qualitative results, comparing our super-resolved outputs against both the original low-resolution and the commercial high-resolution imagery. Finally, we discussed the limitations and potential sources of error in our approach.

\subsection{Dataset and Preprocessing}
Our study focused on the Muzaffarpur district in Bihar, India (centered near 26\textdegree07$'$N, 85\textdegree24$'$E), an agricultural region that experienced severe flooding in October 2022. For this research, we curated the BFCD-22, which contained pairs of satellite images taken before and after the flood event. The dataset consists of 10m resolution of Sentinel-2 scenes and 3m resolution commercial PlanetScope scenes (RGB+NIR).

\begin{figure}[!t]
    \centering
    \includegraphics[width=0.9\textwidth]{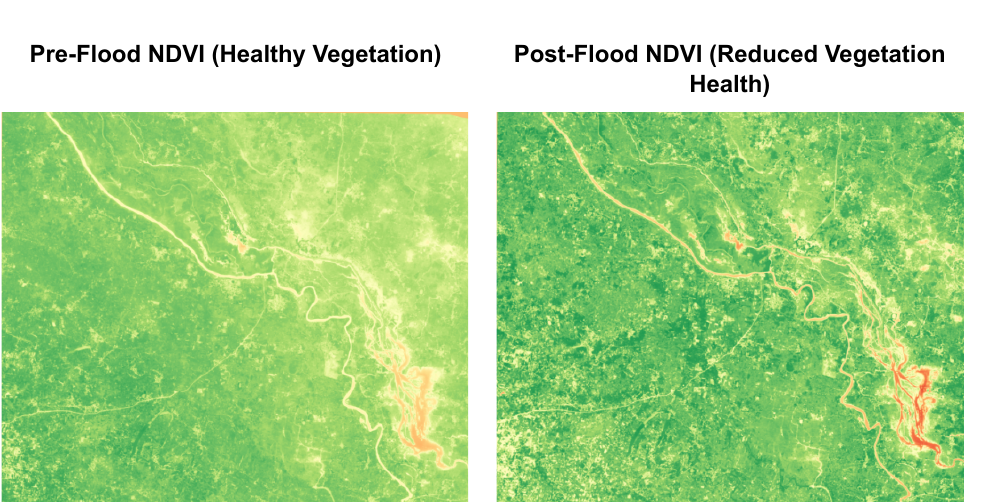}
    \caption{Pre- and post-flood NDVI over the study area in Muzaffarpur (October 2022). Healthy pre-flood vegetation (left) transitions to reduced post-flood vigor (right), motivating $\Delta$NDVI as a damage feature.}
    \label{fig:prepost}
\end{figure}

Our preprocessing workflow was designed to ensure data quality and consistency. First, we applied the official quality masks (L2A for Sentinel-2 and UDM2 for PlanetScope) to remove pixels contaminated by clouds and shadows. All images were then orthorectified and carefully aligned to a common 3m grid through rigid co-registration with manual quality control. This step was critical for accurately comparing pre- and post-flood images. For both satellite sources, we calculated the NDVI using the standard formula \cite{Rouse1974,Tucker1979}:
\begin{equation}
\mathrm{NDVI} \;=\; \frac{\mathrm{NIR}-\mathrm{Red}}{\mathrm{NIR}+\mathrm{Red}}
\end{equation}
Finally, we applied a 10m cropland mask to ensure our analysis was focused strictly on agricultural areas. The final dataset was composed of aligned $256{\times}256$ pixel chips (at the 3m scale) that were ready for training and evaluation. Figure~\ref{fig:prepost} provides a visual overview of the NDVI change in the study area.

\subsection{Implementation Details}
All models were trained and tested on a single NVIDIA A100 40GB GPU, using Python and the PyTorch library. The EDSR model was trained to minimize an $\ell_1$ loss using the Adam optimizer \cite{Kingma2015} with a learning rate of $1{\times}10^{-4}$. We used a batch size of eight and trained for up to one hundred epochs, employing early stopping to prevent overfitting and a learning rate scheduler to adjust the rate during training.

The four-level U\textsc{Net} model was trained to map the single-channel $\Delta$NDVI input to our three damage classes using a cross-entropy loss. We again used the Adam optimizer \cite{Kingma2015} with a learning rate of $1{\times}10^{-3}$, a batch size of eight, and the same early stopping and scheduling strategy.

\subsection{Evaluation Metrics}
To measure the performance of our pipeline, we used two sets of metrics. (a) We evaluated the quality of our super-resolved images using the Peak Signal-to-Noise Ratio (PSNR) and the Structural Similarity Index Measure (SSIM) \cite{Wang2004SSIM}. PSNR measures the pixel-wise reconstruction error, while SSIM measures the similarity in structure between the generated image and the real high-resolution reference. For both metrics, a higher score indicates better performance. The formulas are defined as:
\begin{gather}
\mathrm{MSE}(x,y)=\frac{1}{N}\sum_{i=1}^{N}\big(x_i-y_i\big)^2 \\
\mathrm{PSNR}(x,y)=10\log_{10}\!\left(\frac{L^2}{\mathrm{MSE}(x,y)}\right) \\
\mathrm{SSIM}(x,y)=
\frac{(2\mu_x\mu_y+C_1)(2\sigma_{xy}+C_2)}{(\mu_x^2+\mu_y^2+C_1)(\sigma_x^2+\sigma_y^2+C_2)}
\end{gather}
where $x$ is the reference image, $y$ is the reconstructed image, and $L$ is the dynamic range of the signal (L=2.0 for NDVI).

(b) We evaluated the final damage classification using the class-wise F1-score, which provides a balanced measure of precision and recall for each of the ``No Damage," ``Partial Damage," and ``Full Damage" classes.

\begin{table}[H]
    \centering
    \caption{EDSR performance against 3\,m PlanetScope NDVI.}
    \label{tab:sr_performance}
    \begin{tabular}{@{}lcc@{}}
        \toprule
        \textbf{Event} & \textbf{PSNR} & \textbf{SSIM} \\
        \midrule
        Pre-Flood  & 21.10 & 0.860 \\
        Post-Flood & 20.77 & 0.748 \\
        \bottomrule
    \end{tabular}
\end{table}
\vspace{-1.0cm}

\begin{table}[H]
    \centering
    \caption{F1 score comparison across input modalities.}
    \label{tab:f1_scores}
    \begin{tabular}{@{}lccc@{}}
        \toprule
        \textbf{Source} & \textbf{No Damage} & \textbf{Partial Damage} & \textbf{Full Damage} \\
        \midrule
        Sentinel-2  & 1.00 & 0.98 & 0.83 \\
        PlanetScope & 0.98 & 0.90 & 0.89 \\
        SR (EDSR)   & 0.99 & 0.96 & 0.89 \\
        \bottomrule
    \end{tabular}
\end{table}

\subsection{Quantitative Results}
The performance of our super-resolution model is summarized in Table~\ref{tab:sr_performance}. The model reconstructed the pre-flood NDVI with a high PSNR of 21.10 and SSIM of 0.860, indicating a strong structural alignment with the 3m reference. The post-flood NDVI was more challenging to reconstruct, reflected in slightly lower scores. This was expected due to the chaotic and mixed nature of a landscape immediately after a flood.

\begin{figure}
    \centering
    \includegraphics[width=\textwidth]{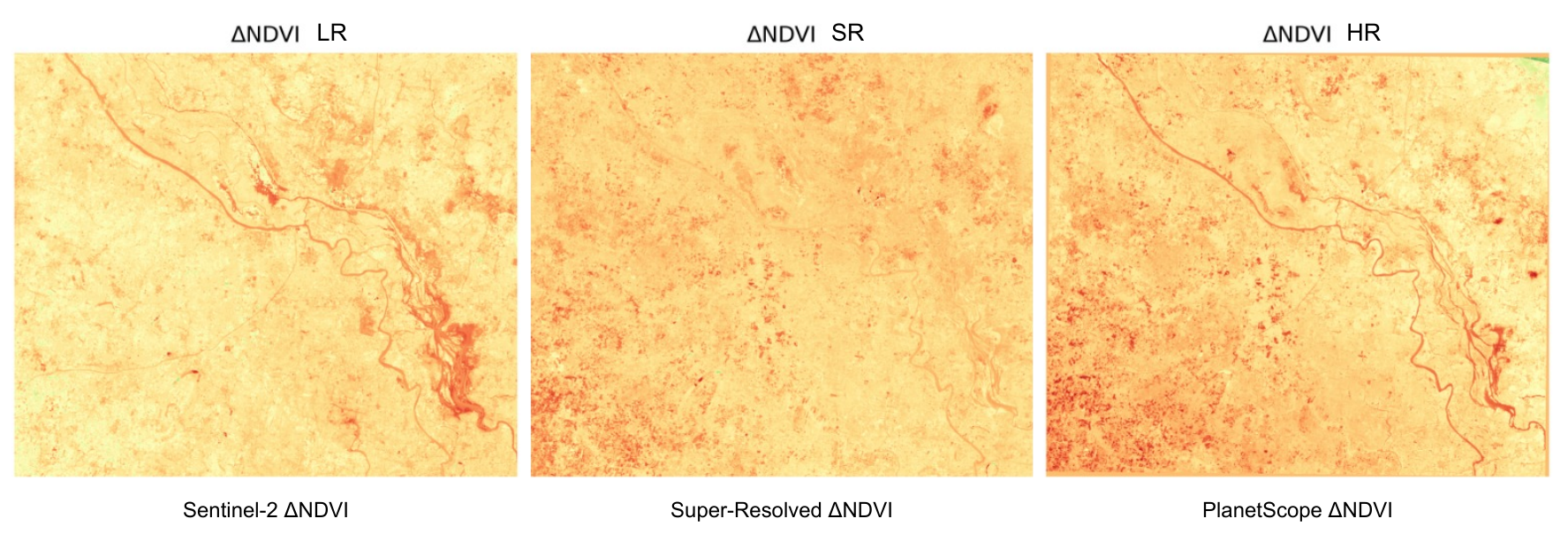}
    \caption{Comparison of $\Delta$NDVI for a representative subregion: (left) native 10\,m Sentinel-2 (LR), (middle) super-resolved 3\,m from Sentinel-2 (SR), (right) native 3\,m PlanetScope (HR). Super-resolution sharpens parcel boundaries and reduces mixed-pixel smearing relative to native 10\,m.}
    \label{fig:comparison}
\end{figure}

\subsection{Qualitative Results}
\begin{figure}
    \centering
    \includegraphics[width=0.9\textwidth]{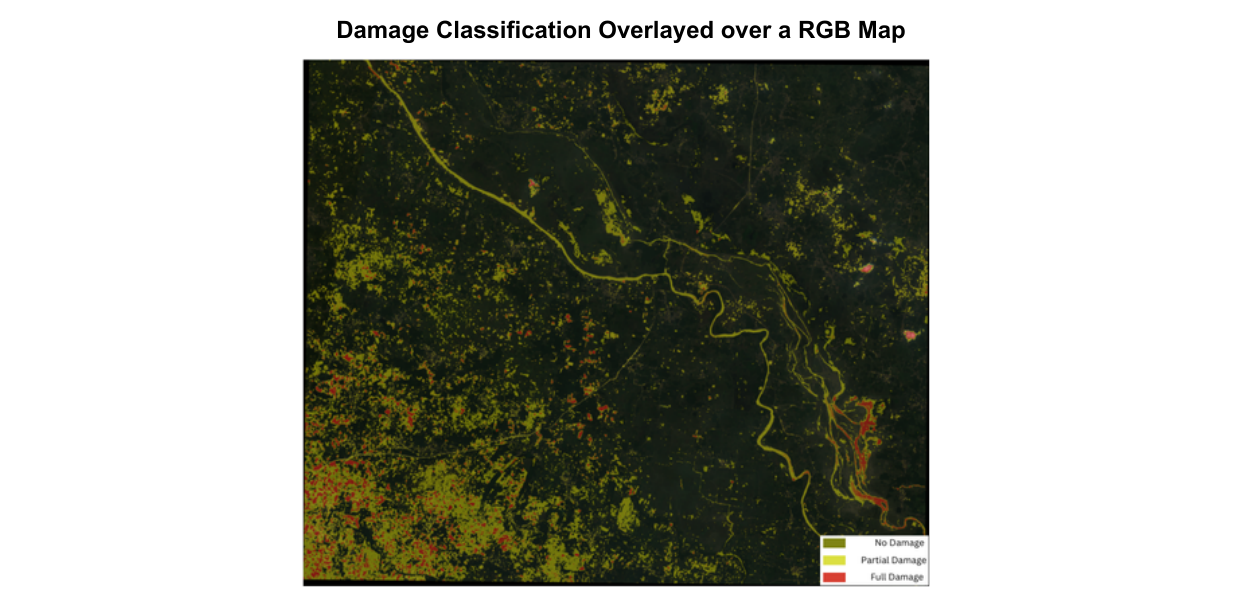}
    \caption{Qualitative output: U\textsc{Net} damage classification overlaid on RGB. Super-resolution sharpening improves boundary fidelity and retrieval of narrow Full-Damage features relative to native 10\,m inputs.}
    \label{fig:damagequal}
\end{figure}

Beyond the quantitative scores, the visual results in Figures~\ref{fig:comparison} and~\ref{fig:damagequal} showed the practical benefit of our method. The super-resolved $\Delta$NDVI map was visibly sharper than the native 10m map, with clearer boundaries and less smearing. This visual improvement aligned with the theoretical benefits of super-resolution and unmixing \cite{Keshava2002,Bioucas2012,Lim2017}. As shown in the figures, our pipeline was better able to identify narrow strips of fully damaged crops, such as those along drainage lines, which were often blurred out in the lower-resolution imagery.

\subsection{Limitations and Error Modes}\label{sec:limits}
While our method showed strong results, it was important to acknowledge its limitations and potential sources of error. Image alignment remains a critical challenge; even tiny sub-pixel misalignments between the pre- and post-flood images can appear as false damage along field edges. Residual contamination from small clouds or water glare that is not caught by the quality masks can also be mistaken for vegetation loss. Furthermore, our threshold-based labeling might occasionally misclassify normal agricultural activities (like harvesting) as damage in some edge cases. Even with super-resolution, very small or narrow parcels can still present a challenge, making it difficult to distinguish between ``Partial" and ``Full" damage \cite{Keshava2002,Bioucas2012}. Finally, this study was conducted for a single event and location; further validation will be needed to test how well the model generalizes to different seasons, crop types, and geographies.

\section{Conclusion}

This paper addressed the critical challenge of accurately assessing flood damage on small farms, where free low-resolution satellite imagery is often too coarse and high-resolution imagery is too expensive. We introduced FLNet, a cost-effective pipeline that used deep learning-based super-resolution to bridge this gap. Our method enhanced 10m Sentinel-2 NDVI data to a 3m resolution using an EDSR model trained on real PlanetScope imagery. A U\textsc{Net} model then classified the change in NDVI ($\Delta$NDVI) into three levels of crop damage. Our experiments on the October 2022 Muzaffarpur flood demonstrated the effectiveness of this approach. The F1-score for the critical "Full Damage" class improved significantly from 0.83 with standard 10m data to 0.89, matching the performance of commercial high-resolution imagery.

In conclusion, our work showed that it is possible to achieve accurate, farm-level damage assessment without relying on costly commercial data. This provided a practical and scalable path forward for developing automated, nationwide systems to support timely flood relief and policy-making. For future work, we will focus on improving the robustness of our pipeline, primarily by finding ways to handle cloud cover. This will involve creating specialized datasets for training cloud removal models \cite{Ghildiyal2025JSTARS} and exploring both optical-only \cite{Ghildiyal2024EJA} and SAR-assisted \cite{Ghildiyal2022MIPR} techniques to reconstruct clear views from cloudy images. Beyond cloud handling, we also plan to investigate architectural improvements, such as directly fusing Sentinel-1 and Sentinel-2 data and experimenting with state-of-the-art backbones like Transformers and Diffusion models.

\section*{Acknowledgments}
Authors acknowledge ANNAM.AI, an AI-CoE of the Ministry of Education, Govt. of India at the Indian Institute of Technology Ropar, for resources and support to execute this work.

\section*{Disclosure of Interests}
The authors have no competing interests to declare that are relevant to the content of this article.

\end{document}